\documentclass[journal,twoside,web]{ieeecolor}
\usepackage{tmi}
\usepackage{colortbl}
\usepackage{cite}
\usepackage{amsmath,amssymb,amsfonts}
\usepackage{algorithmic}
\usepackage{booktabs}
\usepackage{graphicx}
\usepackage{textcomp}
\usepackage{placeins}
\usepackage{float}
\usepackage{stfloats}

\usepackage{multicol}
\usepackage{balance}
\usepackage{lipsum}
\usepackage{multirow}
\usepackage{pifont}
\usepackage{footnote}
\usepackage[dvipsnames]{xcolor}
\let\labelindent\relax
\usepackage{enumitem}
\usepackage{subcaption}
\usepackage[colorlinks=true, linkcolor=Red, citecolor=Green, urlcolor=Blue]{hyperref}
\usepackage{orcidlink}
\usepackage{cite}

\makeatletter
\let\NAT@parse\undefined
\makeatother

\definecolor{mycorrect}{HTML}{82B366}
\definecolor{myerror}{HTML}{B85450}
\definecolor{myhard}{HTML}{D79B00}

\def\BibTeX{{\rm B\kern-.05em{\sc i\kern-.025em b}\kern-.08em
    T\kern-.1667em\lower.7ex\hbox{E}\kern-.125emX}}
\begin{document}
\title{CheXPO-v2: Preference Optimization for Chest X-ray VLMs with Knowledge Graph Consistency}

\author{Xiao Liang\orcidlink{0000-0003-0382-2715}, Yuxuan An, Di Wang\orcidlink{0000-0001-8027-4287}, Jiawei Hu, Zhicheng Jiao\orcidlink{0000-0002-6968-0919}, Bin Jing\orcidlink{0000-0002-4478-8683}, and Quan Wang\orcidlink{0000-0001-6913-8604}
\thanks{This work was supported in part by the National Science and Technology Major Project under Grant 2022ZD0117103, in part by the Outstanding Youth Science Foundation of Shaanxi Province under Grant 2025JC-JCQN-083, in part by the National Natural Science Foundation of China under Grant 62577041, and in part by the Key Research and Development Program of Shaanxi Province under Grant 2025CY-YBXM-047. (Corresponding author: Di Wang.)}
\thanks{X. Liang, Y. An, D. Wang, J. Hu, and Q. Wang are with the School of Computer Science and Technology, Xidian University, Xi'an, China (e-mail: ecoxial2012@outlook.com; wangdi@xidian.edu.cn).}
\thanks{Z. Jiao is with the Warren Alpert Medical School of Brown University, Brown University, Providence, RI, USA.}
\thanks{B. Jing is with the School of Biomedical Engineering, Capital Medical University, Beijing, China.}
\thanks{\textit{This work has been submitted to the IEEE for possible publication. Copyright may be transferred without notice, after which this version may no longer be accessible.}}
}



\maketitle

\begin{abstract}
Medical Vision-Language Models (VLMs) are prone to hallucinations, compromising clinical reliability. While reinforcement learning methods like Group Relative Policy Optimization (GRPO) offer a low-cost alignment solution, their reliance on sparse, outcome-based rewards inadvertently encourages models to ``overthink''---generating verbose, convoluted, and unverifiable Chain-of-Thought reasoning to justify answers. This focus on outcomes obscures factual errors and poses significant safety risks. To address this, we propose \textbf{CheXPO-v2}, a novel alignment framework that shifts from outcome to process supervision. Our core innovation is a Knowledge Graph Consistency Reward mechanism driven by Entity-Relation Matching. By explicitly parsing reasoning steps into structured ``\textit{Disease, Relation, Anatomy}'' triplets, we provide fine-grained supervision that penalizes incoherent logic and hallucinations at the atomic level. Integrating this with a hard-example mining strategy, our approach significantly outperforms GRPO and state-of-the-art models on benchmarks like MIMIC-CXR-VQA. Crucially, CheXPO-v2 achieves new state-of-the-art accuracy using only 5k samples, demonstrating exceptional data efficiency while producing clinically sound and verifiable reasoning. The project source code is publicly available at: \url{https://github.com/ecoxial2007/CheX-Phi4MM}.
\end{abstract}


\section{Introduction}
\label{sec:introduction}

\begin{figure}
    \centering
    \includegraphics[width=1\linewidth]{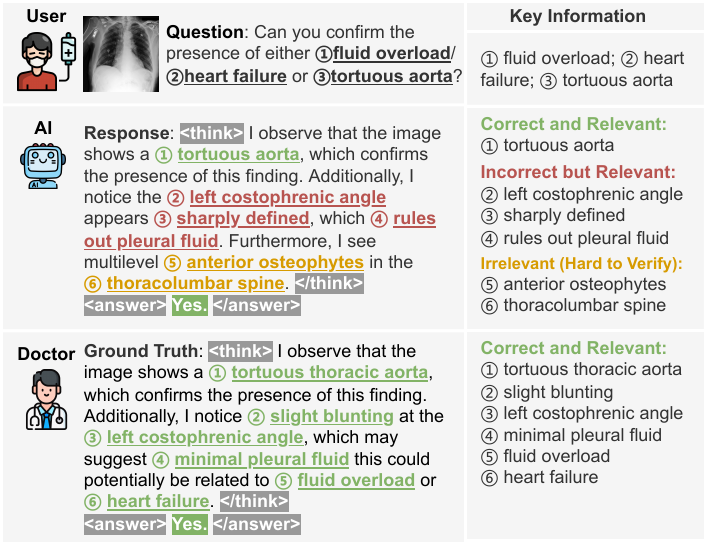}
    \caption{A medical vision-language models's generated response can contain \textcolor{myerror}{factual errors} (e.g., misjudging the angle as \textcolor{myerror}{``sharply defined''}) and \textcolor{myhard}{hard-to-verify}, irrelevant information (e.g., \textcolor{myhard}{``osteophytes''}).}
    \label{fig:motivation}
\end{figure}


\IEEEPARstart{V}{ision}-Language Models (VLMs) have emerged as powerful diagnostic tools, capable of performing complex multimodal reasoning to enhance the accuracy of image analysis~\cite{TMI_MCPL, TMI_VLM_CPL, TMI_Disease_Informed} and report generation~\cite{STREAM_TMI25, MultiPhased_TMI25, dcgrg_liang}. However, a critical flaw inherited from large language models—``\textit{hallucination}''—severely undermines their clinical utility~\cite{Jin2024Hidden}. In a medical context, this flaw manifests as plausible-sounding yet factually incorrect outputs. As illustrated in Figure \ref{fig:motivation}, a model might generate an erroneous reasoning step, such as misidentifying a \textcolor{mycorrect}{``blunted costophrenic angle''} as \textcolor{myerror}{``sharply defined''} and thus incorrectly \textcolor{myerror}{``ruling out pleural fluid''}. It may also include irrelevant or hard-to-verify observations (e.g., \textcolor{myhard}{``anterior osteophytes''}) within its reasoning. Delivered with confident and professional phrasing, such errors are highly deceptive and can directly influence clinical decisions, leading to incorrect diagnoses or inappropriate treatment plans. Even minor inaccuracies erode clinicians' trust, hindering the adoption of AI systems where reliability is paramount~\cite{NatureMed_LLM23, Jin_npj24_GPT4V}. \textit{Consequently, enhancing model factuality and mitigating the risk of hallucination are fundamental prerequisites for the safe and responsible deployment of AI in medicine.}

A promising avenue to address these issues is to refine model behavior with clinical expert feedback during the post-training alignment stage, primarily through Reinforcement Learning from Human Feedback (RLHF)~\cite{Christiano2017DeepRL}. Classic methods like Proximal Policy Optimization (PPO)~\cite{Schulman2017PPO} require training a separate reward model to interpret clinician preferences, whereas Direct Policy Optimization (DPO)~\cite{Rafailov2023DirectPO} offers a more streamlined method that learns directly from preference data. Recently, Group Relative Policy Optimization (GRPO)~\cite{Shao2024DeepSeekMath} has been proposed as an even more direct alignment method. It bypasses the need for meticulously constructed reward models or pairwise preference data, instead leveraging verifiable reward functions to estimate advantage from the relative quality of intra-group responses and guide policy updates. In the medical domain, where expert feedback is costly and labor-intensive, GRPO's direct approach is particularly appealing. However, supervising the model with only rule-based accuracy and format rewards is insufficient and leads to two intertwined problems for clinical applications:

\begin{itemize}
    \item \textbf{Encourages ``Overthinking''}: The model learns to generate verbose and convoluted reasoning chains to reach a correct conclusion. Since the prevailing paradigm~\cite{Shao2024DeepSeekMath, MedVLM_R1_MICCAI} only rewards the final answer, any reasoning path---regardless of its logic or efficiency---is positively reinforced.

    \item \textbf{Creates Safety Risks}: More dangerously, this incentivizes the model to justify a correct answer with flawed or clinically hazardous reasoning. For instance, a model might correctly diagnose a disease based on an incorrect anatomical premise. This hazardous shortcut goes unpenalized as it leads to the correct outcome, posing a severe threat in high-stakes medical scenarios.
\end{itemize}

In this paper, we introduce \textbf{CheXPO-v2}, an advanced \textbf{P}reference \textbf{O}ptimization strategy for \textbf{Che}st \textbf{X}-ray vision-language models, designed to mitigate the critical safety risks of process blindness by shifting from outcome-sparse rewards to automated process supervision. Specifically, we unify multiple benchmarks~\cite{Johnson2019_MIMICCXRAD, Bae2023EHRXQAAM, Hu2023ExpertKI_MedDiffVQA, Wu2021ChestID_ChestImaGenome} into a large-scale visual instruction dataset covering 3 specialized QA tasks and 10 question types, enriched with 640k expert-level reasoning data. Building upon our previous work~\cite{liang2025chexpo}, the proposed CheXPO-v2 establishes a \textbf{Knowledge Graph Consistency Framework} that enhances reliability through three key steps: \noindent \ding{172} Low log-probability instances where the model exhibits low confidence are targeted to sample a high-value subset that effectively exposes reasoning deficits. \noindent \ding{173} A novel reward mechanism driven by \textbf{Entity-Relation Matching} is designed. Unlike black-box approaches, the reasoning chain is explicitly parsed into structured ``\textit{Disease $|$ Relation $|$ Anatomy}'' triplets to provide fine-grained supervision against hallucinatory logic. \noindent \ding{174} This framework is implemented using GRPO to estimate group-relative advantages without a critic model, effectively incentivizing the generation of reasoning chains that are not only accurate in their conclusions but also clinically verifiable.

Our contributions are summarized as follows:
\begin{itemize}
    \item A large-scale chest X-ray vision-language instruction dataset covering basic, region, and comparison tasks, with 10 clinically fine-grained question types and 640k expert-level reasoning chains for supervised fine-tuning.
    
    \item A verifiable preference optimization strategy, \textbf{CheXPO-v2}, which shifts alignment from outcome rewards to process supervision. Integrating hard example mining with Entity-Relation Matching and GRPO, it enforces clinically sound reasoning logic without expensive expert annotations.
    
    \item \textbf{CheXPO-v2} establishes a new SOTA on MIMIC-CXR-VQA and Medical-Diff-VQA. It boosts accuracy and factual consistency using only 5k samples, offering a scalable path toward safer and more interpretable MedVLMs.
\end{itemize}
\section{Related Works}

\subsection{Medical Vision-Language Models}
The development of Vision-Language Models (VLMs) has significantly advanced task-specific applications in the medical field. Using GPT-based instruction generation, Supervised Fine-Tuning (SFT) has become the primary method for adapting general-purpose VLMs to meet medical needs. Early adaptations like LLaVA-Med \cite{llava-med} employed alignment and instruction tuning to enable open-ended question answering for biomedical images, while Med-Flamingo \cite{pmlr-v225-moor23aMed-Flamingo} enhanced few-shot reasoning through a multimodal context learning framework. Recent works have scaled this approach: RadFM \cite{DBLP:journals/corr/abs-2308-02463_RadFM} integrated large-scale radiology datasets for robust multi-task analysis, and HuatuoVision \cite{chen2024huatuogptvisioninjectingmedicalvisual} generated 1.3 million VQA instances to train high-quality QA models.

To address the domain gap and data scarcity, recent research has shifted toward more sophisticated adaptation strategies beyond standard SFT. For instance, MCPL \cite{TMI_MCPL} introduces anatomical-pathological prompt learning with a graph-guided collaborative module to align text and images. Similarly, \cite{TMI_Knowledge_Intervention} proposes medical knowledge intervention prompt tuning to inject LLM knowledge into the visual encoding process. Addressing the challenge of novel pathologies, \cite{TMI_Disease_Informed} utilizes disease-informed adaptation with prototype learning to recognize unseen diseases from few samples. In the domain of pathology, VLM-CPL \cite{TMI_VLM_CPL} and MSCPT \cite{TMI_MSCPT} leverage VLMs for annotation-free classification and few-shot Whole Slide Image (WSI) analysis, respectively, utilizing multi-scale prompt tuning. Furthermore, to enhance interpretability and segmentation, \cite{TMI_Gaze_Insight} bridges human visual attention with VLM explanations, while AMVLM \cite{TMI_AMVLM} employs alignment-multiplicity awareness for semi-supervised segmentation. Despite these advancements, SFT-based methods inherently focus on next-token prediction, a task biased toward memorizing training patterns. This limitation often results in hallucinations and limits adaptability in complex clinical scenarios where data distributions are long-tailed or contain erroneous entries.

\subsection{Verifiable Reinforcement Learning for Vision-Language Models}
To overcome the limitations of SFT, Reinforcement Learning (RL) has emerged as a critical paradigm for enhancing model reliability and reasoning. While classical RLHF and Direct Preference Optimization (DPO) \cite{Rafailov2023DirectPO} have been widely used to align models with human preferences \cite{sun2024stllava, Zhu2024MMedPOAM, liang2025chexpo}, they often rely on opaque reward models or noisy preference pairs. Recent advancements have pivoted toward Verifiable Reinforcement Learning, specifically leveraging Group Relative Policy Optimization (GRPO) with rule-based or verifiable rewards to incentivize genuine reasoning capabilities (often termed ``R1-style'' reasoning). In the general domain, R1-VL \cite{ICCV25_R1_VL} and Visual-RFT \cite{ICCV25_Visual_RFT} have demonstrated that defining visual verifiable rewards (e.g., IoU, accuracy) can significantly boost performance in fine-grained classification and grounding. Additionally, DocThinker \cite{ICCV25_DocThinker} and Table-R1 \cite{Table_R1} utilize rule-based rewards to enforce structural logic in document and table reasoning tasks.

In the medical domain, this paradigm is rapidly gaining traction. Med-R1 \cite{Med_R1_2025} and MedVLM-R1 \cite{MedVLM_R1_MICCAI} introduce RL frameworks to incentivize medical reasoning, shifting focus from pattern matching to logical deduction. For specific tasks, MedGround-R1 \cite{MedGround_R1} utilizes spatial-semantic rewards for grounding, while Efficient Medical VIE \cite{Medical_VIE} applies GRPO to information extraction with precision-recall balanced rewards. PathVLM-R1 \cite{PathVLM_R1} further extends this to pathology tasks. However, current GRPO approaches face stability and consistency challenges. \cite{Toward_Effective_RL} analyzes the impact of initialization and answer length in medical GRPO, while Hint-GRPO \cite{Hint_GRPO} proposes text-debiased hints for hard examples. Most relevant to our work are approaches addressing reasoning consistency; GRPO-CARE \cite{GRPO_CARE} introduces a consistency-aware reward to ensure the Chain-of-Thought (CoT) aligns with the final answer. Unlike these methods, our proposed framework specifically targets the clinical factual consistency of the reasoning process itself, utilizing fine-grained entity-relation extraction to penalize medical hallucinations at the atomic level.


\section{Data Synthesis and Model Warm-up}
\label{sec:data}
To investigate the process of injecting specialized medical knowledge into a general-purpose large model from the ground up, and subsequently enhancing its clinical reasoning capabilities through reinforcement learning, we developed a comprehensive data synthesis pipeline and a tailored model warm-up strategy. This section details our approach to generating a high-quality chest X-ray instruction dataset and the supervised fine-tuning methodology used to prepare the model.

\subsection{Synthetic Instruction Data for Chest X-ray Analysis}
\label{sec:sft_data}
Our data synthesis pipeline begins with the consolidation of four key public datasets: MIMIC-CXR \cite{Johnson2019_MIMICCXRAD} for images and reports, Chest ImaGenome \cite{Wu2021ChestID_ChestImaGenome} for expert-labeled anatomy, and both MIMIC-CXR-VQA \cite{Bae2023EHRXQAAM} and Medical-Diff-VQA \cite{Hu2023ExpertKI_MedDiffVQA} for question-answer pairs. A critical limitation of existing VQA datasets \cite{Bae2023EHRXQAAM, Hu2023ExpertKI_MedDiffVQA} is the absence of intermediate Chain-of-Thought (CoT) steps, which restricts models from learning verifiable diagnostic logic. To address this, we leverage GPT-4o \cite{openai2024gpt4o} to synthesize comprehensive reasoning chains. By conditioning on questions, answers, and radiology reports, we generate a high-quality instruction set simulating expert clinical reasoning across three tasks:

\begin{figure}[htp]
    \centering
    \includegraphics[width=1\linewidth]{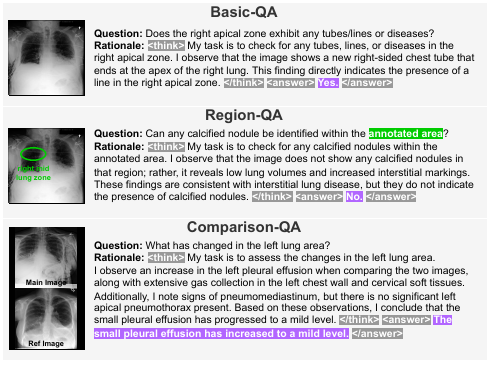}
    \caption{
\textbf{Examples from the multi-task QA dataset.} Each entry includes a question $\mathcal{Q}$ and a ground-truth reference $\mathcal{R}$. This reference contains a Chain-of-Thought (CoT) $\mathcal{T}$ enclosed by \texttt{<think>} tags, followed by the final answer $\mathcal{A}$ within \texttt{<answer>} tags. The {\colorbox[HTML]{00CC00}{\textcolor{white}{``annotation area''}}} highlights the visual prompt on the image.
    }
    \label{fig:dataset_case}
\end{figure}

\begin{table}[t]
  \centering
  \caption{\textbf{Question Types and Examples.}}
  \label{tab:question_types_examples}
  \resizebox{\columnwidth}{!}{
    \begin{tabular}{lp{25em}}
    \toprule
    \textbf{Question Type} & \textbf{Example} \\
    \midrule
    Presence & Is any technical assessment present within the left breast? \\
    Anatomy & Enumerate all the regions in the anatomy that are related to vascular congestion. \\
    Attribute & Provide a list of all the tubes/lines found in both the carina and neck. \\
    Abnormality & Can you identify any abnormalities? \\
    Size  & Is the cardiac silhouette wider than 50\% of the thorax's width? \\
    Plane & Does this image have an AP projection? \\
    Gender & Is the patient identified as male? \\
    Severity & What level is the pleural effusion? \\
    Type  & What type is the edema? \\
    Difference & What has changed compared to the reference image? \\
    \bottomrule
    \end{tabular}%
  }
\end{table}

\begin{table*}[t]
  \centering
  \caption{\textbf{Dataset statistics.} ``Closed'' refers to yes/no questions, while ``Open'' includes other answer types. }
  \label{tab:dataset}%
    \begin{tabular}{l|l|l|rrrr}
    \toprule
    \textbf{Subtask} & \multicolumn{1}{l|}{\textbf{Anno. Source}} & \textbf{Split} & \multicolumn{1}{l}{\textbf{QA Pairs}} & \multicolumn{1}{l}{\textbf{Closed}} & \multicolumn{1}{l}{\textbf{Open}} & \multicolumn{1}{l}{\textbf{Images}} \\
    \midrule
    \multirow{3}[2]{*}{\textbf{Basic}} & \multicolumn{1}{p{8.465em}|}{MIMIC-CXR-VQA} & Train & 181,349 & 117,394 & 63,955 & 91,406 \\
          & \multicolumn{1}{l|}{MIMIC-CXR} & Valid & 44,264 & 28,597 & 15,667 & 6,201 \\
          &       & Test  & 9,066 & 5,704 & 3,362 & 409 \\
    \midrule
    \multirow{3}[2]{*}{\textbf{Region}} & \multicolumn{1}{l|}{MIMIC-CXR-VQA} & Train & 44,686 & 34,064 & 10,622 & 36,293 \\
          & \multicolumn{1}{l|}{MIMIC-CXR} & Valid & 10,574 & 8,061 & 2,513 & 5,009 \\
          & \multicolumn{1}{l|}{Chest ImaGenome} & Test  & 2,134 & 1,605 & 529   & 408 \\
    \midrule
    \multirow{3}[2]{*}{\textbf{Compare}} & \multicolumn{1}{l|}{Medical-Diff-VQA} & Train & 319,859 & 165,089 & 154,770 & 110,761 \\
          & \multicolumn{1}{l|}{MIMIC-CXR} & Valid & 14,176 & 7,202 & 6,974 & 5,827 \\
          &       & Test  & 13,646 & 6,901 & 6,745 & 5,706 \\
    \bottomrule
    \end{tabular}%

\end{table*}%

\textbf{Basic-QA}. We utilize structured prompting templates that integrate existing question-answer pairs from MIMIC-CXR-VQA \cite{Bae2023EHRXQAAM} and corresponding radiology reports from MIMIC-CXR \cite{Johnson2019_MIMICCXRAD}. The model is instructed to refer to these reports and infer supporting evidence from the image to justify the given answer. To enforce a structured CoT format, GPT-4o is prompted to generate a reasoning block \(\mathcal{T}\) enclosed within \texttt{<think>} tags, followed immediately by the final answer \(\mathcal{A}\) inside \texttt{<answer>} tags. We explicitly instruct the model to adopt an expert radiologist's perspective, ensuring \(\mathcal{T}\) provides a logical derivation of the gold-standard answer for the question \(\mathcal{Q}\).

\textbf{Region-QA}. To enhance interpretability, we build on existing question-answer pairs by adding explicit visual prompts \textbf{via direct overlays}, and we replace specific anatomical terms in the question with placeholders such as \textit{``marked region''} or \textit{``highlighted area.''} The visual prompts refer to the bounding box coordinates provided by the Chest ImaGenome \cite{Wu2021ChestID_ChestImaGenome} silver dataset. To further enhance spatial flexibility, we additionally incorporate prompt types such as arrows, ellipses, or asterisks (*), following the approach of \cite{Cai2023ViPLLaVAML}. The arrowheads are positioned within a constrained region $[(-w/2, -h/2),$ $ (w/2, h/2)]$, where \(w\) and \(h\) represent the bounding box width and height, respectively. For ellipses, the axes inherit dimensions from the bounding box and are scaled within a range of \([1, 1.5]\). These augmentations are inspired by visual prompts commonly seen in PubMed figures, where experts directly overlay markers on images for clearer illustration, providing a more intuitive and flexible approach.

\textbf{Comparison-QA}. Medical-Diff-VQA\cite{Hu2023ExpertKI_MedDiffVQA} provides study IDs for both the main and reference chest X-rays being compared. We leverage these IDs to retrieve their respective radiology reports, using them as contrastive evidence. GPT-4o is then prompted to generate detailed comparative reasoning chains based on the main report and the prior report. Similar to Basic and Region-QA, we instruct GPT-4o to emulate expert image analysis, requiring the generated CoT to derive exclusively from visual evidence while strictly avoiding references to textual reports or prior diagnostic annotations.

We unify the question types from MIMIC-CXR-VQA and Medical-Diff-VQA into 10 clinically relevant categories, as detailed in Table \ref{tab:question_types_examples}. Figure \ref{fig:dataset_case} presents representative examples across the three tasks, showcasing the structured CoT reasoning. To ensure fair benchmarking, we follow the original train/validation/test splits, with additional cleaning to address missing MIMIC-CXR reports, absent Chest ImaGenome bounding boxes, and overlapping images across benchmarks. The final dataset statistics are summarized in Table \ref{tab:dataset}. All the above datasets are publicly available on the PhysioNet platform \cite{PhysioNet} and require credentialed access approval.

\subsection{Model Warm-up via Supervised Fine-Tuning}
\label{sec:warmup}

The primary goal of the warm-up phase is to adapt the pre-trained model to the medical domain and our specific instruction format. Inspired by the DeepSeek-R1 cold-start strategy \cite{Shao2024DeepSeekMath}, this is achieved via Supervised Fine-Tuning (SFT) on a synthesized dataset following the CoT template. To address the lack of native reasoning tokens in the base Phi-4 models \cite{Abouelenin2025Phi4MiniTR}, we first extend the vocabulary by adding four special tokens—\texttt{<think>}, \texttt{</think>}, \texttt{<answer>}, and \texttt{</answer>}—to the tokenizer and resizing the model's token embedding layer. The model is then fine-tuned using Low-Rank Adaptation (LoRA) \cite{Hu2021LoRALA}, with the \textit{lm\_head} and the newly added token embeddings unfrozen for training. This warm-up phase efficiently teaches the model both domain knowledge and the required output format simultaneously, significantly accelerating convergence.

\begin{figure*}[t]
    \centering
    \includegraphics[width=1\linewidth]{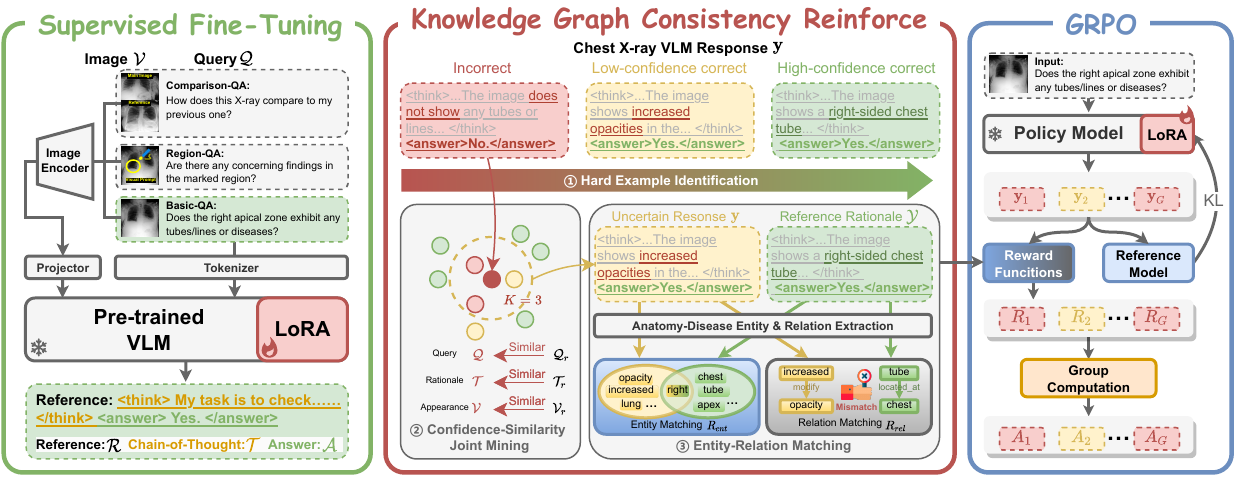}
    \caption{\textbf{Overview of our Knowledge Graph Consistency Framework.} This two-stage pipeline enhances MedVLM reasoning via Group-Relative Policy Optimization. High-value training examples are selected using \textit{Hard Example Mining}. The reward signal is determined by the \textit{Knowledge Graph Consistency Reward}, which uses Anatomy-Disease Entity and Relation Extraction to provide fine-grained process supervision.}
    \label{fig:framework}
\end{figure*}

\section{Methodology}
\label{sec:method}

Although the Supervised Fine-Tuning (SFT) phase enables the model to learn reasoning formats and exposure to medical images, it can cause the model to overfit to the head of the training distribution. Consequently, the model struggles with long-tail cases and is more prone to hallucination when faced with out-of-distribution scenarios \cite{Huang2023ASO, chusft}. To overcome this limitation, we employ the Group Relative Policy Optimization (GRPO) algorithm \cite{Shao2024DeepSeekMath} via reinforcement learning to enhance the model's reasoning capabilities. In the following sections, we will first introduce the preliminaries of GRPO and then detail our proposed process reward function, with the overall framework outlined in Figure~\ref{fig:framework}.

\subsection{Preliminaries: Group Relative Policy Optimization}
\label{sec:preliminaries_grpo}
GRPO \cite{Shao2024DeepSeekMath} enhances reasoning by optimizing a policy based on the relative advantages of a group of candidate outputs. Unlike conventional RL methods such as PPO \cite{Schulman2017PPO}, GRPO bypasses the need for a separate, learnable function to estimate state-action values. Instead, it calculates the advantage for each candidate response by comparing its reward to the mean and standard deviation of the entire group of candidates generated for the same prompt. At each training step, for a given input $q$, GRPO first samples a group of $G$ candidate outputs $\{o_i\}_{i=1}^G$ from the current policy $\pi_{\theta_{old}}$, which is initialized via SFT. A reward $R_i$ is computed for each output $o_i$, and the group-relative advantage $A_i$ is then calculated by normalizing the rewards within the group.
\begin{equation}
\label{equ:advantage}
    A_i = \frac{R_i - \text{mean}(\{R_1, ..., R_G\})}{\text{std}(\{R_1, ..., R_G\})}.
\end{equation}

This formulation ensures that outputs with above-average rewards are positively reinforced. The policy $\pi_\theta$ is then updated by maximizing the GRPO objective function, which incorporates a clipped importance sampling ratio to ensure training stability:

\begin{equation}
\label{eq:grpo}
\begin{aligned}
\mathcal{L}_{\mathrm{GRPO}}(\theta) & = \mathbb{E}_{\substack{q \sim P(Q), \{o_{i}\}_{i=1}^{G} \sim \pi_{\theta_{old}}}(\cdot \mid q)} \\
& \quad \frac{1}{G} \sum_{i=1}^{G} 
\Biggl[
    \min 
        \left(
            \frac{\pi_{\theta_{\text{new}}}(o_{i} \mid q)}{\pi_{\theta_{old}}(o_{i} \mid q)} A_{i}, 
        \right. 
\Biggr. 
\\
& \quad \operatorname{clip} 
\left.
    \left( 
        \frac{\pi_{\theta_{\text{new}}}(o_{i} \mid q)}{\pi_{\theta_{old}}(o_{i} \mid q)}, 1-\varepsilon, 1+\varepsilon_{high} 
    \right) A_{i} 
\right) 
\\
& \quad - 
\Biggl. 
    \beta \mathbb{D}_{\mathrm{KL}} 
        \left(
            \pi_{\theta_{\text{new}}} \| \pi_{ref}
        \right)
\Biggr].
\end{aligned}
\end{equation}

The Kullback-Leibler (KL) divergence term is included to prevent catastrophic forgetting by penalizing significant deviations from the reference model $\pi_{ref}$. Its specific formulation is as follows:

\begin{equation}
\label{eq:kl_divergence}
\mathbb{D}_{KL}\left(\pi_{\theta_{\text{new}}} \| \pi_{r e f}\right)=\frac{\pi_{r e f}\left(o_{i} \mid q\right)}{\pi_{\theta_{\text{new}}}\left(o_{i} \mid q\right)}-\log \frac{\pi_{r e f}\left(o_{i} \mid q\right)}{\pi_{\theta_{\text{new}}}\left(o_{i} \mid q\right)}-1.
\end{equation}

The group-relative advantage $A_i$ in the foundational GRPO method \cite{Shao2024DeepSeekMath} is computed from a simple, rule-based reward signal. Under this scheme, a response is rewarded based on the correctness of its final answer and its adherence to the required output format. However, this reward mechanism suffers from process blindness; it evaluates only the final outcome and format, remaining ignorant of the logical quality and factual accuracy of the internal reasoning steps.

\subsection{Hard Example Mining}
\label{sec:hard_example}

While GRPO effectively improves model generalization, its requirement of sampling \(G\) responses per prompt incurs a high computational cost compared to SFT. To enhance training efficiency, we follow our previous work~\cite{liang2025chexpo} and introduce a hard example mining strategy that identifies high-value data for GRPO training, specifically targeting cases where the base SFT model fails. Given a chest X-ray vision-language model $\pi_{\theta}$ trained via supervised fine-tuning and an SFT training dataset $\mathcal{D}$ consisting of images $\mathcal{V}$, questions $\mathcal{Q}$, and ground-truth references $\mathcal{R}$, the model generates responses $\mathbf{y}$ structured as $[\mathbf{y}_\mathcal{T}; \mathbf{y}_\mathcal{A}]$, which consist of a reasoning chain $\mathbf{y}_\mathcal{T}$ followed by the final answer $\mathbf{y}_\mathcal{A}$.

To efficiently evaluate model performance and identify hard examples, we perform stratified random sampling from the training dataset based on question and answer types and obtain a representative subset \(\mathcal{D}_{sample}\) for initial assessment. The stratified sampling ratio, denoted as \(\gamma\), represents the proportion of the dataset selected, with \(\gamma < 5\%\) of the total training data, ensuring comprehensive coverage of all question types. Within this subset, for each image $\mathcal{V}$ and corresponding question $\mathcal{Q}$, the model $\pi_{\theta}$ generates a token sequence \(\mathbf{y}\) in the CoT format, structured as $[\mathbf{y}_\mathcal{T}; \mathbf{y}_\mathcal{A}]$. As only the short answer segments ($\mathbf{y}_\mathcal{A}$) have readily verifiable ground-truth labels, we focus our evaluation on them. To score a generated answer, we compute its length-normalized log-probability under the model's policy ($\pi_{\theta}$), a technique inspired by \cite{Malinin2021UncertaintyEI}:

\begin{equation}
\label{eq:length_norm_policy}
\mathbf{p}(\mathbf{y}_\mathcal{A}) = \frac{1}{|\mathbf{y}_\mathcal{A}|} \sum_{t=1}^{|\mathbf{y}_\mathcal{A}|} \log \pi_{\theta}(y_t \mid y_{<t}, \mathcal{V}, \mathcal{Q}, \mathbf{y}_\mathcal{T}).
\end{equation}

When $\mathbf{y}_\mathcal{A} \neq \mathcal{A}$, the data sample $\{\mathcal{V}, \mathcal{Q}, \mathcal{R}\}$ is used as a high-value training sample for GRPO to induce policy updates against similar errors. This provides a low answer reward $R$ to significantly update the model's policy and reduce the probability of making similar mistakes. For cases where $\mathbf{y}_\mathcal{A} = \mathcal{A}$, it is difficult to directly assess the quality of the response \(\mathbf{y}\). We thus select samples where the model's confidence $\mathbf{p}$ is low (i.e., $\mathbf{p} < \sigma$) as hard examples with high entropy. These low-confidence, correct samples likely yield a high variance in rewards across the generated group, providing a rich and effective gradient for the GRPO's advantage-based objective function, which improves model robustness and expedites convergence \cite{Hint_GRPO}.

After initializing the hard example set \(\mathcal{D}_{hard}\), we retrieve semantically similar samples from the remaining training data \(\mathcal{D}_{rest} = \mathcal{D} - \mathcal{D}_{hard}\) to further enrich the dataset for GRPO training, thereby mitigating SFT failures caused by the long-tailed distribution. We first leverage BioMedCLIP ~\cite{Zhang2023BiomedCLIPAM} to extract features $f_\mathcal{V}$, $f_\mathcal{Q}$, and $f_\mathcal{T}$ for the images, questions, and rationale in $\mathcal{D}$. Then, we compute cosine similarity matrices $\mathcal{S}$ between $\mathcal{D}_{hard}$ and $\mathcal{D}_{rest}$ for $f_\mathcal{Q}$, $f_\mathcal{T}$, and $f_\mathcal{V}$, respectively, to identify samples with similar semantics and visual context:

\begin{equation}
    \mathcal{S}_\mathcal{X} = \Big[ \langle f_\mathcal{X}^{(i)}, f_\mathcal{X}^{(j)} \rangle \Big]_{\substack{i \in \mathcal{D}_{hard} \\ j \in \mathcal{D}_{rest}}}, \quad \mathcal{X} \in \{\mathcal{Q}, \mathcal{T}, \mathcal{V}\},
\end{equation} 
where $\langle \cdot, \cdot \rangle$ denotes cosine similarity. Final similarity combines all modalities: 
\begin{equation}
\label{eq:combined_simi}
\mathcal{S} = \mathcal{S}_\mathcal{Q} + \mathcal{S}_\mathcal{T} + \mathcal{S}_\mathcal{V},
\end{equation} 
where $\mathcal{S} = \{\mathbf{s}^{(i,j)}\}_{i \in \mathcal{D}_{hard}, j \in \mathcal{D}_{rest}}$. We then select the Top-$K$ most similar samples $\mathcal{N}^{(i)}$ for $i$-th hard example based on $\mathcal{S}$:
\begin{equation}
\label{eq:topk}
\mathcal{N}^{(i)} = \text{TopK}_{j \in \mathcal{D}_{rest}} \, \mathcal{S}^{(i,j)}, \quad \forall i \in \mathcal{D}_{hard},
\end{equation} 
and this combined similarity ensures a balance between semantic alignment and visual consistency. Subsequently, the retrieved Top-$K$ samples are aggregated into the hard example set $\mathcal{D}_{hard}$ for GRPO training, where each training instance is formalized as a triplet $\{\mathcal{V}, \mathcal{Q}, \mathcal{R}\}$ containing the image, question, and ground-truth reference.

\subsection{Knowledge Graph Consistency Reward}
\label{sec:reward_function}
To guide the model towards generating clinically sound reasoning, we introduce a multi-faceted reward function. This function evaluates the model's generated response \(\mathbf{y}_\mathcal{T}\) by decomposing it into structured knowledge and validating it against the corresponding ground-truth reasoning \(\mathcal{T}\). The process involves extracting entities and their relations from the free-text reasoning within the \texttt{<think>}... \texttt{</think>} block before calculating a consistency-aware reward.

\subsubsection{Anatomy-Disease Entity Extraction}
First, we parse the generated reasoning text \(\mathbf{y}_\mathcal{T}\) to identify key medical entities. Based on a schema tailored for radiology, we extract mentions of six entity types: \textit{Anatomy}, \textit{Disorder}, \textit{Concept} (e.g., ``acute'', ``increased''), \textit{Device} (e.g., ``tube'', ``clip''), \textit{Procedure}, and \textit{Size}. To perform this extraction, we leverage the specialized Named Entity Recognition (NER) model trained via ReXKG~\cite{ReXKG_zhang2024uncovering}. For a given reasoning sequence, this process yields a set of unique entities, which we denote as \(\mathcal{E}\).

\subsubsection{Anatomy-Disease Relation Extraction}
Following entity extraction, we identify the relationships between them to form a structured understanding of the reasoning. We define three primary directed relations:
\begin{itemize}
    \item \textbf{Located at:} A source entity is located at a target entity (e.g., \textit{tube} located at \textit{chest}).
    \item \textbf{Suggestive of:} A source entity (e.g., a finding) suggests a target entity (e.g., a disease).
    \item \textbf{Modify:} A source entity provides additional information about a target entity (e.g., \textit{increased} modify \textit{opacity}).
\end{itemize}

A relation extraction model processes the entities $\mathcal{E}$ to produce a set of relation triplets, denoted as $\mathcal{K} = \{(e_h, r, e_t)\}$, where $e_h, e_t \in \mathcal{E}$ are the head and tail entities, and $r$ is one of the three relation types.

\subsubsection{Entity-Relation Matching}
The total reward signal is a composite of scores evaluating the knowledge graph consistency of the reasoning process, measured via Entity-Relation Matching, and the correctness of the final answer.

\textbf{Entity Matching ($R_{ent}$):} This reward measures the entity matching between the entity sets extracted from the generated reasoning $\mathbf{y}_\mathcal{T}$ and the ground-truth reasoning $\mathcal{T}$, denoted as $\mathcal{E}_{\mathbf{y}_\mathcal{T}}$ and $\mathcal{E}_{\mathcal{T}}$, respectively. We define it as the Jaccard index between these sets, which penalizes hallucinations and omissions:
\begin{equation}
\label{eq:ent_reward}
        R_{ent} = \frac{|\mathcal{E}_{\mathbf{y}_\mathcal{T}} \cap \mathcal{E}_{\mathcal{T}}|}{|\mathcal{E}_{\mathbf{y}_\mathcal{T}} \cup \mathcal{E}_{\mathcal{T}}|}.
\end{equation}

\textbf{Relation Matching ($R_{rel}$):} Similarly, this reward assesses the relation matching between extracted triplets. It is defined as the Jaccard of ground-truth relations, ensuring that the model forms logically and factually correct connections between entities.
\begin{equation}
\label{eq:rel_reward}
    R_{rel} = \frac{|\mathcal{K}_{\mathbf{y}_\mathcal{T}} \cap \mathcal{K}_{\mathcal{T}}|}{|\mathcal{K}_{\mathbf{y}_\mathcal{T}} \cup \mathcal{K}_{\mathcal{T}}|}.
\end{equation}

\textbf{Answer Reward:} Given that the SFT phase has largely resolved issues related to the required output format (e.g., ensuring the presence of \texttt{<answer>}... \texttt{</answer>} blocks), we primarily focus on rewarding the correctness of the final answer matching the ground truth $\mathcal{A}$. This component follows the original implementation of GRPO \cite{Shao2024DeepSeekMath}, which is designed to ensure the model produces the correct answer within the designated block.

Finally, the total reward \(R\) is computed as a weighted sum: \(R = w_1 R_{ans} + w_2 R_{ent} + w_3 R_{rel}\), where we set \(w_1=1\), \(w_2=0.5\), and \(w_3=0.5\). This reward is subsequently utilized to estimate the group-relative advantage in Equation~\ref{equ:advantage} and guide the policy optimization process defined in Equation~\ref{eq:grpo}.

\section{Experiments}
\label{sec:experiments}

\begin{table}[htbp]
  \centering
  \caption{\textbf{Hyperparameter settings.}}
  \label{tab:training_parameter}
    \begin{tabular}{llcc}
    \toprule
    \multicolumn{2}{l}{\textbf{Hyper-Parameter}} & \textbf{SFT} & \textbf{RL} \\ \midrule
    \multirow{4}[0]{*}{\textbf{Optimizer}} & batch size & 64    & 8 \\
          & epoch & 1 & 1 \\
          & learning rate & 6e-4 & 4e-6 \\
          & warmup ratio & 0.05  & 0.1 \\ \midrule
    \multirow{2}[0]{*}{\textbf{Model}} & max\_new\_tokens & \multicolumn{2}{c}{256} \\
          & num\_crop & \multicolumn{2}{c}{4} \\ \midrule
    \multirow{5}[1]{*}{\textbf{LoRA}} & lora\_rank & \multicolumn{2}{c}{256} \\
          & lora\_alpha & \multicolumn{2}{c}{512} \\
          & \multirow{3}[1]{*}{target\_modules} & \multicolumn{2}{c}{qkv\_proj, o\_proj, } \\
          &       & \multicolumn{2}{c}{gate\_up\_proj,} \\
          &       & \multicolumn{2}{c}{down\_proj} \\
    \midrule

    \multirow{4}[1]{*}{\textbf{RL Policy}} & beta $\beta$  & \multirow{4}[1]{*}{——} & 0.5 \\
          & epsilon $\epsilon$ &       & 0.2 \\
          & epsilon\_high $\epsilon_{high}$ &       & 0.28 \\
          & num\_generation $G$ &       & 8 \\ \midrule
    \end{tabular}

\end{table}%

\subsection{Implementation Details}
\label{sec:implementation}

We fine-tune the Phi-4-multimodal (Phi-4MM) \cite{Abouelenin2025Phi4MiniTR} model using LoRA \cite{Hu2021LoRALA}, updating only the injected layers. Our training pipeline consists of two stages conducted on different hardware configurations due to computational requirements: the efficient SFT stage is performed on an NVIDIA RTX 4090 (24GB), while the GRPO stage, which requires generating multiple candidate responses, utilizes an NVIDIA A100 (80GB). Detailed hyperparameters are listed in Table~\ref{tab:training_parameter}. Both stages employ the AdamW optimizer with a cosine scheduler. Specifically, the SFT stage uses a learning rate of 6e-4 and a batch size of 64 for 1 epoch. The GRPO stage uses a learning rate of 4e-6, a batch size of 8, and $\beta = 0.5$. Regarding data sampling, we employ a stratified sampling strategy to avoid the computational cost of forwarding the entire dataset. We uniformly sample with a ratio $\gamma = 2\%$, resulting in $\sim$10k samples for initial assessment. With a log-probability threshold $\sigma = -0.25$ (prob $\approx 0.78$), we identify $\sim$3k high-value samples for training, supplemented by retrieval-based augmentation \cite{liang2025chexpo} when scaling is needed.

\subsection{Evaluation Metrics}
\label{sec:metrics}

To thoroughly evaluate both the final output and the fidelity of the reasoning process, we employ a multi-dimensional assessment framework consisting of three components: 

\subsubsection{Answer Accuracy}
Answer accuracy verifies the final diagnostic conclusion. This metric compares the content generated within the \texttt{<answer>}\dots\texttt{</answer>} block against the ground-truth reference using strict string matching.

\subsubsection{Entity-Relation Matching}
To assess the quality of the generated reasoning process $\mathbf{y}_\mathcal{T}$, we extract structured knowledge elements, specifically entities $\mathcal{E}$ and relations $\mathcal{K}$, from the \texttt{<think>}\dots\texttt{</think>} blocks. We compare these extracted items against the ground-truth reasoning $\mathcal{T}$ via Entity-Relation Matching to report Precision, Recall, and F1 score. These metrics quantify the model's ability to accurately and comprehensively capture relevant clinical details during inference.

\subsubsection{Knowledge Graph Structural Metrics}
We construct knowledge graphs $KG_{\mathcal{T}}$ and $KG_{\mathbf{y}_\mathcal{T}}$ based on the extracted entities and relations from the ground truth and generated reasoning, respectively. Adopting the ReXKG framework~\cite{ReXKG_zhang2024uncovering}, we introduce three complementary metrics:

\textbf{KG Node Similarity Coefficient (KG-NSC).} This metric quantifies entity coverage by identifying the most semantically similar node in $KG_{\mathbf{y}_\mathcal{T}}$ for each node in $KG_{\mathcal{T}}$ and calculating the average of these maximum similarity scores:
\begin{equation}
    \text{KG-NSC} = \frac{1}{N} \sum_{i=1}^{N} s_i,
\end{equation}
where $N$ is the number of nodes in $KG_{\mathcal{T}}$ and $s_i$ is the maximum similarity score for the $i$-th ground-truth node.

\textbf{KG Adjacency Matrix Similarity (KG-AMS).} To evaluate relationship consistency, we map nodes from $KG_{\mathbf{y}_\mathcal{T}}$ to $KG_{\mathcal{T}}$ to construct two aligned adjacency matrices, denoted as $M_{\mathbf{y}_\mathcal{T}}$ and $M_{\mathcal{T}}$. We measure the structural similarity by calculating the weighted Pearson correlation coefficient between the corresponding rows of these matrices:
\begin{equation}
    \text{KG-AMS} = \frac{\sum_{i=1}^{N} w_{r_i} \cdot \text{corr}(M_{\mathbf{y}_\mathcal{T},i}, M_{\mathcal{T},i})}{\sum_{i=1}^{N} w_{r_i}},
\end{equation}
where $M_{\mathcal{T},i}$ and $M_{\mathbf{y}_\mathcal{T},i}$ represent the $i$-th row of the ground-truth and predicted adjacency matrices, respectively, and $w_{r_i}$ denotes the weight of the $i$-th row.

\textbf{KG Subgraph Coverage Score (KG-SCS).} This metric assesses the reproduction of salient clinical findings, represented as critical subgraphs (e.g., \textit{Disorder}-\textbf{Located at}-\textit{Anatomy}). We calculate the weighted average of presence scores for the top-$K$ important subgraphs in $KG_{\mathcal{T}}$:
\begin{equation}
    \text{KG-SCS} = \frac{\sum_{i=1}^{K} I(S_i) P(S_i)}{\sum_{i=1}^{K} I(S_i)},
\end{equation}
where $I(S_i)$ denotes the importance score of subgraph $S_i$. The presence score $P(S_i)$ combines edge and node matching:
\begin{equation}
    P(S_i) = \frac{1}{2} \left( \frac{|E(\hat{S}_i)|}{|E(S_i)|} + \frac{\sum_{n \in N(S_i)} s_n}{|N(S_i)|} \right),
\end{equation}
where $\hat{S}_i$ is the corresponding subgraph in $KG_{\mathbf{y}_\mathcal{T}}$, and $s_n$ is the node similarity score derived from KG-NSC.

\section{Results and Analysis}
\label{sec:results}

\begin{table*}[t]
  \centering
  \caption{\textbf{Overall accuracy (\%) on chest X-ray VQA.} Benchmarks include supervised fine-tuning performance using Short Answer (SA) vs. Chain-of-Thought (CoT) formats on Phi-3.5V and Phi-4MM models. Furthermore, our prior CheXPO \cite{liang2025chexpo} using counterfactual DPO is compared against the proposed CheXPO-v2 using knowledge graph consistency GRPO, which sets a new state-of-the-art with only 5k samples.}
    
  \label{tab:sota-results}
  \resizebox{\linewidth}{!}{
    \begin{tabular}{lcccccccccccccc}
    \cmidrule{1-11}\cmidrule{13-15}    \textbf{Model} & \multicolumn{10}{c}{\textbf{Question Type}}                                   &       & \multicolumn{2}{c}{\textbf{Answer Type}} & \textbf{Overall} \\
    \cmidrule{2-11}\cmidrule{13-14}          & Presence & Abnormality & Anatomy & Severity & Plane & Type  & Difference & Attribute & Size  & Gender &       & Open  & Closed &  \\
        \midrule
        \rowcolor[rgb]{ .851,  .851,  .851} \multicolumn{15}{l}{\textit{General Vision Language Models}} \\
        LLaVA-v1.6-7B & 67.67  & 63.97  & 17.92  & 0.00  & 67.71  & 23.81  & 0.00  & 15.17  & 58.18  & 70.00  &       & 11.07  & 77.52  & 47.26  \\
        Qwen2.5VL-Chat-7B & 66.95  & 63.02  & 21.67  & 18.81  & 88.95  & 14.29  & 81.25  & 21.38  & 45.45  & 36.67  &       & 25.91  & 76.31  & 53.36  \\
        Phi3.5V-4.2B & 57.96  & 27.47  & 11.19  & 1.11  & 38.03  & 0.00  & 0.00  & 17.24  & 18.18  & 63.33  &       & 7.47  & 51.19  & 31.30  \\
        Phi4MM-6B & 51.70  & 29.63  & 23.64  & 0.00  & 35.00  & 0.00  & 66.67  & 27.87  & 38.46  & 66.67  &       & 21.62  & 47.88  & 38.17  \\
        \midrule
        \rowcolor[rgb]{ .851,  .851,  .851} \multicolumn{15}{l}{\textit{Medical Visual Language Models}} \\
        Med-Flamingo & 62.45  & 57.67  & 17.36  & 14.16  & 50.99  & 12.38  & 0.00  & 17.93  & 65.45  & 50.00  &       & 14.84  & 67.72  & 43.64  \\
        RadFM & 59.51  & 48.53  & 12.36  & 7.08  & 40.79  & 8.57  & 0.00  & 13.79  & 63.64  & 63.33  &       & 9.75  & 62.06  & 38.24  \\
        LLaVA-Med-7B & 68.21  & 64.22  & 17.92  & 0.00  & 68.56  & 27.14  & 0.00  & 16.55  & 45.45  & 70.00  &       & 11.51  & 77.78  & 47.60  \\
        HuatuoV-7B & 76.42  & 63.88  & 22.36  & 32.96  & 83.00  & 20.00  & 48.08  & 24.83  & 83.64  & 53.33  &       & 26.17  & 83.07  & 57.16  \\
        HuatuoV-34B & 90.82  & 61.55  & 22.92  & 54.87  & 89.24  & 14.76  & 0.00  & 27.59  & \textbf{92.73 } & 66.67  &       & 27.27  & 90.97  & 61.96  \\
        \midrule
        \rowcolor[rgb]{ .851,  .851,  .851} \multicolumn{15}{l}{\textit{Baseline Implementation}} \\
        Phi3.5V+SA & 81.00  & 62.89  & 44.87  & 60.03  & 95.69  & 63.84  & 91.18  & 73.21  & 63.45  & 66.98  &       & 55.82  & 81.76  & 69.95  \\
        Phi3.5V+CoT & 86.26  & 71.46  & 43.86  & 62.16  & 98.30  & 64.74  & 91.18  & 78.80  & 63.62  & 70.68  &       & 57.53  & 87.77  & 73.20  \\
        Phi4MM+SA & 76.23  & 64.44  & 60.00  & 36.00  & 95.00  & 63.64  & 100.00  & 67.21  & 53.85  & \textbf{100.00}  &       & 59.46 & 76.19  & 70.00  \\
        Phi4MM+CoT & 79.25  & 71.11  & \underline{70.91}  & 44.00  & 98.87  & 63.64  & 100.00  & 72.13  & 76.92  & 83.33  &       & 66.67  & 80.16  & 75.17  \\
        \midrule
        \rowcolor[rgb]{ .851,  .851,  .851} \multicolumn{15}{l}{\textit{Our Proposal}} \\
        CheXPO-Phi3.5V(20k) & 89.55  & 76.84  & 46.40  & \underline{67.10}  & 98.58  & \underline{65.90}  & 91.13  & \underline{80.88}  & 66.38  & 73.77  &       & 60.87  & 91.10  & 77.34  \\
        CheXPO-Phi3.5V(30k) & \underline{91.39}  & 80.68  & 49.35  & \textbf{69.12 } & 98.87  & 65.72  & 91.18  & \textbf{90.51 } & 66.03  & 76.54  &       & 63.12  & \underline{93.64}  & 79.74  \\
        CheXPO-v2-Phi4MM(1k) & 87.55  & \underline{82.96}  & \textbf{78.18 } & 52.00  & \textbf{100.00 } & 54.55  & \textbf{100.00 } & 73.77  & \underline{92.31} & 83.33  &       & \textbf{70.72 } & 89.95  & \underline{82.80}  \\
        CheXPO-v2-Phi4MM(5k) & \textbf{92.07 } & \textbf{85.93 } & \underline{76.36}  & 52.00  & \textbf{100.00 } & \textbf{72.73 } & \textbf{100.00 } & 75.41  & \underline{92.31} & \textbf{100.00 } &       & \textbf{70.72 } & \textbf{94.97 } & \textbf{86.00 } \\
        \bottomrule
        \end{tabular}%
    
    } 

\end{table*}%

\subsection{Comparison with State-of-the-Art Models}
\label{sec:sota_comparison}

We compare our method against two categories of vision-language models: 1) \textbf{General-purpose VLMs}, including LLaVA-v1.6 \cite{llava}, Qwen-VL-Chat \cite{qwen}, Phi-3.5V \cite{Abdin2024Phi3TR}, and Phi-4MM \cite{Abouelenin2025Phi4MiniTR}. 2) \textbf{Medical VLMs}, including Med-Flamingo \cite{pmlr-v225-moor23aMed-Flamingo}, RadFM \cite{DBLP:journals/corr/abs-2308-02463_RadFM}, LLaVA-Med \cite{llava-med}, and HuatuoVision \cite{chen2024huatuogptvisioninjectingmedicalvisual}. To ensure consistent answer formatting, we provide each model with one-shot exemplars selected according to the question-answer type. We also fine-tune Phi-3.5V and Phi-4MM on our dataset using two rationale settings: short answer (SA) and Chain-of-Thought (CoT). The results are summarized in Table~\ref{tab:sota-results}. Our SFT notably improves performance over the baselines, raising Phi-3.5V from 31.30\% to 73.20\% and Phi-4MM from 38.17\% to 75.17\%. Crucially, the CoT setting consistently outperforms the SA setting, demonstrating the value of reasoning traces. Finally, compared to our previous work \cite{liang2025chexpo}, our advanced CheXPO-v2 strategy demonstrates significantly higher data efficiency and performance, achieving a superior state-of-the-art accuracy of 86.00\% using only 5k preference samples versus the 30k required previously.

\subsection{Ablation Studies}
\label{sec:ablation}
In this section, we conduct a series of ablation studies to systematically dissect the contributions of our key design choices to the performance of MedVLM. Our investigation is twofold: 1) we explore how different data sampling strategies, particularly \textbf{Hard Example Mining}, during the SFT and RL phases impact the model's reasoning abilities; 2) we analyze the individual and combined effects of our \textbf{Knowledge Graph Consistency Reward}. These experiments provide critical insights into the optimal configuration for training clinically sound and reliable reasoning models.

\begin{figure*}[!t]
    \centering
    \includegraphics[width=1\textwidth]{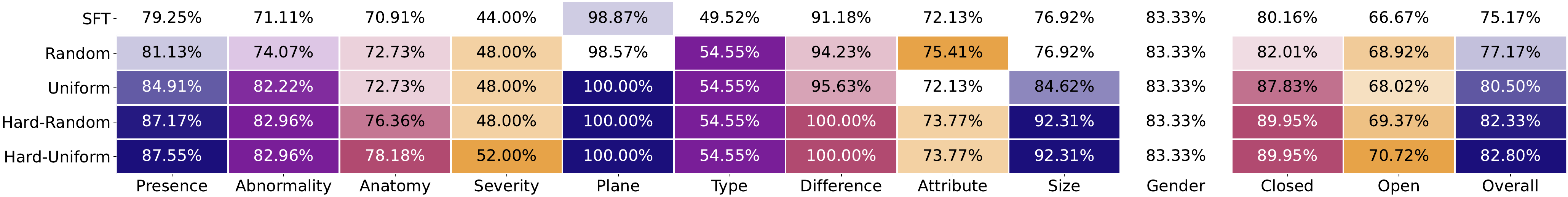}
    \vspace{-0.5cm}
    \caption{\textbf{Type-wise accuracy (\%) on chest X-ray VQA across different sampling strategies.} The default setting uses 1k samples and a composite reward of answer correctness and Entity-Relation Matching (Jaccard).}
    \label{fig:ablation_data_sample}
\end{figure*}

\begin{figure}[htbp]
\centering
\begin{minipage}[b]{0.235\textwidth}
    \centering
    \includegraphics[width=\textwidth]{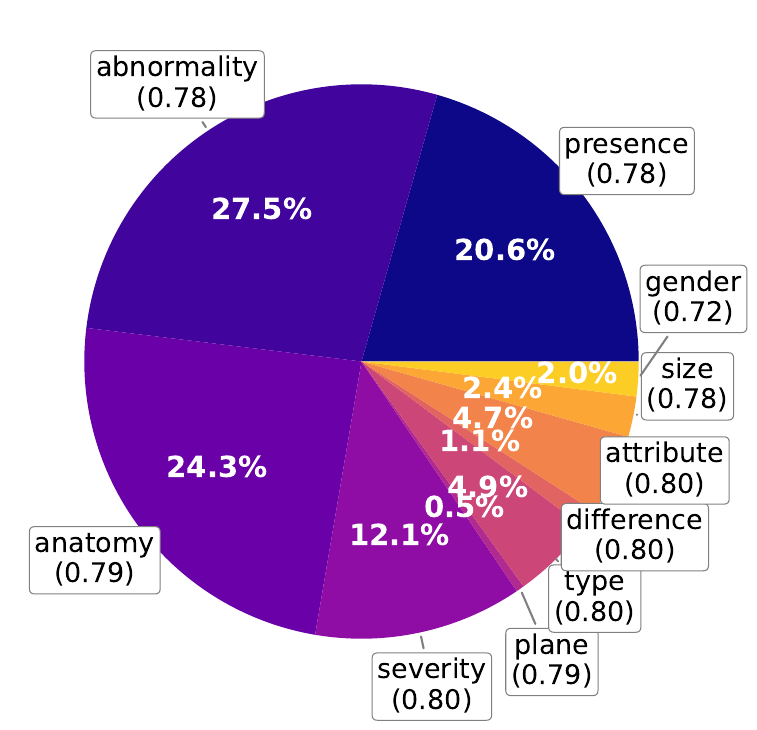}
    \vspace{-5pt} 
    \centerline{\small \textbf{(a)} }
\end{minipage}
\hfill
\begin{minipage}[b]{0.235\textwidth}
    \centering
    \includegraphics[width=\textwidth]{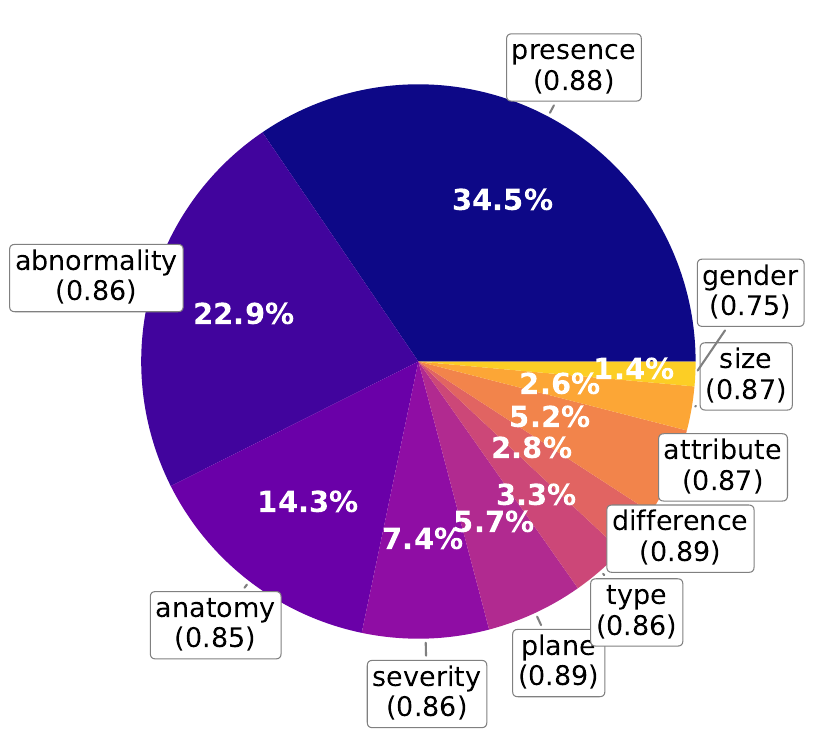}
    \vspace{-5pt} 
    \centerline{\small \textbf{(b)} }
\end{minipage}
\caption[Hard example distribution analysis]{ \textbf{Comparison of distribution between SFT failure cases (a) and the original dataset (b), with the average token-level probability of the content between \texttt{<answer>} and \texttt{</answer>}.} }
\label{fig:class_distribution}
\end{figure}

\begin{figure}[htbp]
\centering  
\includegraphics[width=1\columnwidth]{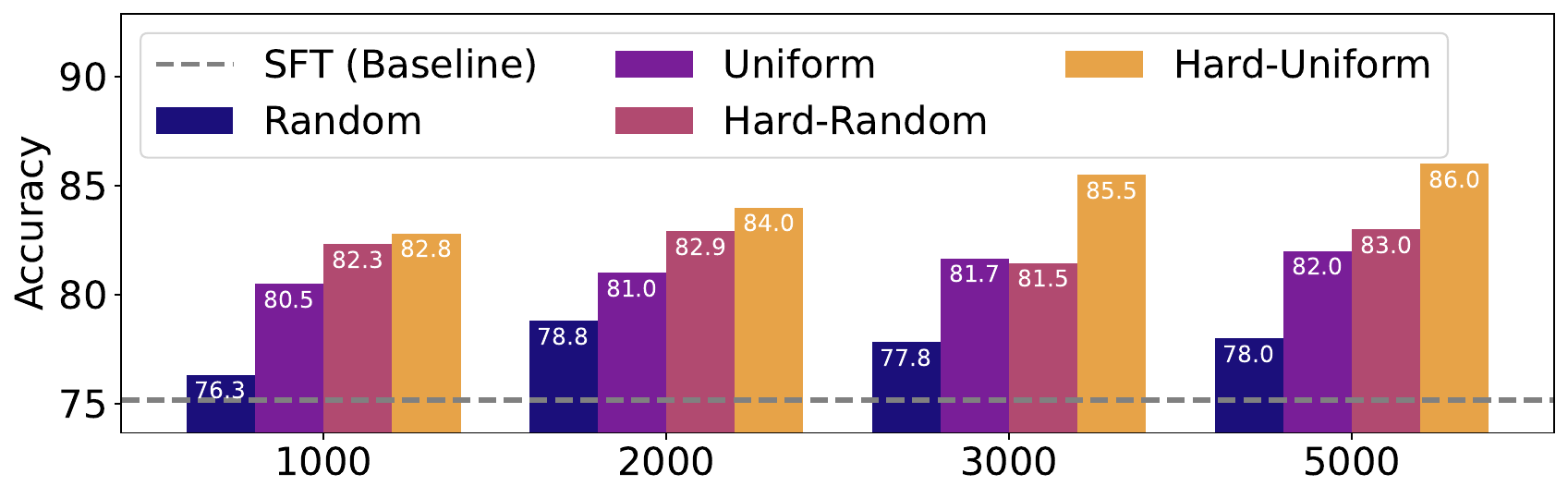} 
\vspace{-0.5cm}
\caption{ \textbf{Impact of sampling strategies and preference dataset size on diagnostic accuracy.} We compare different sampling methods across varying data scales, optimizing with the default reward setting.}
\label{fig:data_scaling}  
\end{figure}

\subsubsection{Impact of Data Sampling Strategies}
\label{sec:ablation_data}

We evaluate the impact of different data curation strategies on model performance, proceeding from the analysis of failure cases to the comparison of sampling methods and data scaling efficiency.

\textbf{Necessity of Hard Example Mining.} 
Figure \ref{fig:class_distribution} compares the failure case distribution of the SFT model against the overall test set distribution across the \textbf{Basic}, \textbf{Region}, and \textbf{Comparison} QA subsets. We also analyze the average token-level probability of the answer \(\mathcal{A}\). Observations indicate that \textit{Abnormality}, \textit{Anatomy}, and \textit{Severity} dominate the failure cases—accounting for 27.5\%, 24.3\%, and 12.1\% respectively—which contrasts with their lower prevalence of 22.9\%, 14.3\%, and 7.4\% in the original distribution. Specifically, \textit{Abnormality} covers anatomical findings, diseases, and devices; \textit{Anatomy} involves open-ended location queries; and \textit{Severity} assesses symptom levels. Most failure cases exhibit lower probabilities, such as 0.78 for \textit{Abnormality} compared to the category average of 0.86, whereas simple tasks like \textit{Plane} determination remain highly reliable. Without targeted hard mining, the vast majority of correct samples exhibit high probability and low entropy, providing negligible reward signals for policy optimization. This results in low data efficiency, as the model focuses on maintaining performance on easy samples rather than correcting high-stakes errors in complex diagnostic scenarios.

\begin{table*}[ht]
  \centering
  \caption{\textbf{Ablation study of reward function components.} We compare baselines against variants utilizing different knowledge graph contents and similarity metrics. Results report answer accuracy, entity-relation consistency, knowledge graph structural metrics, and CoT length statistics.}
  \label{tab:kg_ablation}%
  \resizebox{\linewidth}{!}{
    \begin{tabular}{lcccccccccccccc}
    \toprule
    \textbf{Reward Function} & \multicolumn{2}{c}{\textbf{Answer}} &       & \multicolumn{7}{c}{\textbf{Reasoning Fidelity}}        &       & \multicolumn{3}{c}{\textbf{CoT Length}} \\
\cmidrule{2-3}\cmidrule{5-11}\cmidrule{13-15}          & \textbf{Acc(↑)} & \textbf{Hit Rate(↑)} &       & \textbf{Precision(↑)} & \textbf{Recall(↑)} & \textbf{F1(↑)} &       & \textbf{KG-NSC(↑)} & \textbf{KG-AMS(↑)} & \textbf{KG-SCS(↑)} &       & \textbf{min} & \textbf{avg} & \textbf{max} \\
    \midrule
    \rowcolor[rgb]{ .847,  .847,  .847} \multicolumn{15}{l}{\textit{(a) Baseline}} \\
    SFT   & 0.752 & 0.437 &       & 0.514 & 0.410 & 0.440 &       & 0.825 & 0.711 & 0.783 &       & 33.8  & 56.7  & 108.2 \\
    w/ acc & 0.805 & 0.483 &       & 0.503 & 0.410 & 0.437 &       & 0.821 & 0.721 & 0.797 &       & 36.5  & 65.6  & 130.4 \\
    \midrule
    \rowcolor[rgb]{ .847,  .847,  .847} \multicolumn{15}{l}{\textit{(b) Ablating KG Content}} \\
    w/ Jaccard (E) & 0.763 & 0.477 &       & 0.511 & 0.408 & 0.437 &       & 0.840 & 0.734 & 0.803 &       & 33.8  & 57.1  & 115.5 \\
    w/ Jaccard (R) & 0.767 & 0.468 &       & 0.519 & 0.413 & 0.446 &       & 0.825 & 0.702 & 0.847 &       & 35.4  & 60.2  & 128.4 \\
    w/ Jaccard (E+R) & \textbf{0.778} & \textbf{0.478} &       & \textbf{0.520} & \textbf{0.421} & \textbf{0.449} &       & 0.830 & 0.719 & 0.852 &       & 35.9  & 60.0  & 121.5 \\ \midrule
    \rowcolor[rgb]{ .847,  .847,  .847} \textit{(c) Ablating KG Metric} &       &       &       &       &       &       &       &       &       &       &       &       &       &  \\
    w/ acc + F1 (E+R) & 0.802 & 0.488 &       & 0.517 & 0.414 & 0.444 &       & 0.821 & 0.705 & 0.899 &       & 35.9  & 60.2  & 125.0 \\
    w/ acc + Precision (E+R) & 0.807 & 0.492 &       & \textbf{0.540} & 0.399 & 0.443 &       & 0.830 & 0.719 & 0.852 &       & 33.2  & 58.2  & 124.6 \\
    w/ acc + Recall (E+R) & 0.788 & 0.477 &       & 0.507 & 0.417 & 0.442 &       & 0.822 & 0.725 & 0.871 &       & 35.6  & 60.4  & 126.6 \\
    \textbf{w/ acc + Jaccard (E+R)} & \textbf{0.828} & \textbf{0.502} &       & 0.522 & \textbf{0.421} & \textbf{0.450} &       & \textbf{0.880} & \textbf{0.781} & \textbf{0.904} &       & 35.7  & 61.3  & 126.2 \\
    \bottomrule
    \end{tabular}%
    }

\end{table*}%

\begin{table*}[htbp!]
  \centering
  \caption{\textbf{GPT-4o assessment of reasoning fidelity.} We prompt GPT-4o to evaluate the semantic quality by comparing the generated reasoning $\mathbf{y}_{\mathcal{T}}$ directly against the ground truth $\mathcal{T}$.}
  \label{tab:addlabel}%
  \resizebox{\linewidth}{!}{
       \begin{tabular}{lccccccccc}
    \toprule
    \textbf{Reward Function} & \multicolumn{4}{c}{\textbf{Answer Correct}} &       & \multicolumn{4}{c}{\textbf{Answer Error}} \\
\cmidrule{2-5}\cmidrule{7-10}          & \textbf{Overall(↑)} & \multicolumn{3}{c}{\textbf{Reasoning Semantic Fidelity}} &       & \textbf{Overall(↓)} & \multicolumn{3}{c}{\textbf{Reasoning Semantic Fidelity}} \\
\cmidrule{3-5}\cmidrule{8-10}          &       & \multicolumn{1}{p{4.925em}}{\textbf{Correct(↑)}} & \multicolumn{1}{p{5.925em}}{\textbf{Part Error(↓)}} & \multicolumn{1}{p{3.845em}}{\textbf{Error(↓)}} &       &       & \multicolumn{1}{p{4.925em}}{\textbf{Correct(↑)}} & \multicolumn{1}{p{5.925em}}{\textbf{Part Error(↓)}} & \multicolumn{1}{p{3.845em}}{\textbf{Error(↓)}} \\
    \midrule
    \rowcolor[rgb]{ .847,  .847,  .847} \multicolumn{10}{l}{\textit{(a) Baseline}} \\
    SFT   & 0.752 & 0.440 & 0.235 & 0.077 &       & 0.248 & 0.002 & 0.245 & 0.002 \\
    w/ acc & 0.805 & 0.415 & 0.280 & 0.110 &       & 0.195 & 0.005 & 0.187 & 0.003 \\
    \midrule
    \rowcolor[rgb]{ .847,  .847,  .847} \multicolumn{10}{l}{\textit{(b) Ablating KG Content}} \\
    w/ Jaccard(E) & 0.763 & 0.465 & 0.223 & \textbf{0.075} &       & 0.237 & 0.005 & 0.232 & 0.000 \\
    w/ Jaccard(R) & 0.767 & 0.460 & \textbf{0.217} & 0.090 &       & 0.233 & \textbf{0.007} & 0.227 & 0.000 \\
    w/ Jaccard(E+R) & \textbf{0.778} & \textbf{0.470} & 0.222 & 0.087 &       & \textbf{0.222} & 0.005 & \textbf{0.217} & 0.000 \\
    \midrule
    \rowcolor[rgb]{ .847,  .847,  .847} \multicolumn{10}{l}{\textit{(c) Ablating KG Metric}} \\
    w/ acc + F1(E+R) & 0.802 & 0.450 & 0.270 & \textbf{0.082} &       & 0.198 & 0.003 & 0.193 & 0.002 \\
    w/ acc + Precision(E+R) & 0.807 & 0.423 & 0.293 & 0.090 &       & 0.193 & \textbf{0.007} & 0.187 & \textbf{0.000} \\
    w/ acc + Recall(E+R) & 0.788 & 0.480 & \textbf{0.222} & 0.087 &       & 0.212 & 0.005 & 0.203 & 0.003 \\
    \textbf{w/ acc + Jaccard(E+R)} & \textbf{0.828} & \textbf{0.487} & 0.257 & 0.085 &       & \textbf{0.172} & \textbf{0.007} & \textbf{0.165} & \textbf{0.000} \\
    \bottomrule
    \end{tabular}
    }

\end{table*}%

\textbf{Effectiveness of Uniform Hard Mining.} 
Based on the identified hard examples, we evaluate different sampling strategies using a fixed budget of 1k samples. Figure \ref{fig:ablation_data_sample} presents the overall and type-wise accuracy. The \textit{SFT} baseline achieves an overall accuracy of 75.17\%. Simple \textit{Random} sampling yields a modest improvement to 77.17\%. \textit{Uniform} sampling, which balances question types, further improves performance to 80.50\%, validating the benefit of distribution balancing. Notably, our proposed \textit{Hard-Uniform} strategy achieves the best overall performance of 82.80\%. It specifically boosts performance in the most challenging categories identified above, such as \textit{Anatomy} improving from 70.91\% to 78.18\% and \textit{Abnormality} rising from 71.11\% to 82.96\%. This confirms that combining hard example mining with uniform distribution alignment provides the most effective gradients for GRPO optimization.

\textbf{Data Efficiency and Scalability.} 
Finally, we investigate how these improvements scale with dataset size. Figure \ref{fig:data_scaling} compares the accuracy trajectories of different strategies as the preference data increases from 1k to 5k. While hard example mining generally proves effective, strategies relying on random sampling exhibit instability and fail to guarantee consistent performance growth. In contrast, the \textit{Hard-Uniform} strategy consistently outperforms other methods across all data scales. Remarkably, \textit{Hard-Uniform} achieves 82.8\% accuracy with only 1k samples, surpassing the \textit{Random} sampling baseline which reaches approximately 82\% even with 5k samples. As the data size increases to 5k, the \textit{Hard-Uniform} strategy reaches a peak accuracy of 86.0\%, significantly widening the gap with the SFT baseline. This indicates that our approach is not only efficient in low-data regimes but also scales effectively to further enhance model reliability.

\begin{figure*}[h!]
\centering  
\includegraphics[width=1.5\columnwidth]{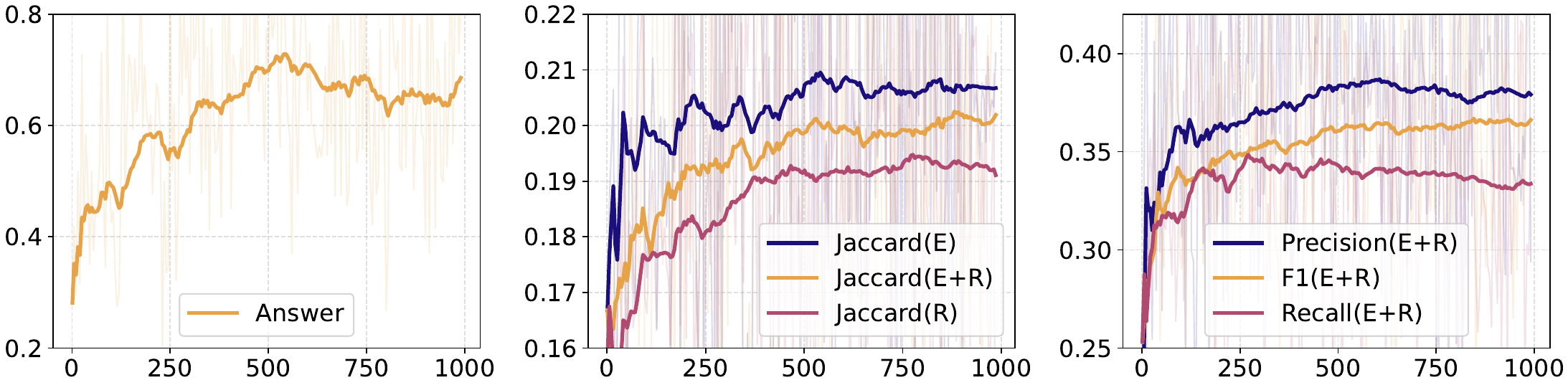}  
\caption{\textbf{Training reward trajectories under the 1k Hard-Uniform setting.} Evolution of the answer reward, Entity-Relation Matching (Jaccard), and fine-grained reasoning metrics (Precision, Recall, F1) over training steps.}
\label{fig:reward_traj}  
\end{figure*}

\subsubsection{Impact of Reward Function Components}
\label{sec:ablation_reward}

We analyze the contribution of each component within our composite reward function. We start with a baseline trained only on the accuracy reward \(R_{\text{acc}}\) and systematically introduce Entity \(R_{\text{ent}}\) and Relation \(R_{\text{rel}}\) consistency rewards.

\textbf{Content of Reward.}
We first investigate the impact of rewarding different reasoning contents including entities and relations. In addition to the metrics introduced in Subsection \ref{sec:metrics}, we also report the Hit Rate metric, which measures the proportion of generated answers appearing directly within the \texttt{<think>} block. Table~\ref{tab:kg_ablation} (b) shows that incentivizing Entity (E) or Relation (R) consistency alone yields performance gains over the SFT baseline. Notably, the Jaccard E+R setting achieves 0.778 accuracy and a 0.478 hit rate. This indicates that the model learns to infer the answer within the reasoning chain without explicit answer supervision. Table~\ref{tab:addlabel} (b) further validates this using GPT-4o assessment. While the accuracy-only reward in part (a) improves final answers, it results in a lower reasoning fidelity of 0.415 compared to 0.470 for Jaccard E+R. Furthermore, the exclusive reliance on outcome supervision degrades reasoning quality, evidenced by the drop in CoT Semantic Fidelity correct rate from 0.440 to 0.415, and promotes verbose generation according to the CoT Length statistics in Table~\ref{tab:kg_ablation}. Conversely, process rewards provide fine-grained signals to ensure logical soundness and conciseness.

\textbf{Metric of Reward.}
We further evaluate different similarity metrics for calculating the process reward including F1, Precision, Recall, and Jaccard. Table~\ref{tab:kg_ablation} (c) demonstrates that the Jaccard similarity metric achieves the superior overall accuracy of 0.828 and the highest KG-NSC score of 0.880. While the Precision-based reward yields the highest precision of 0.540, it sacrifices recall. The Jaccard metric effectively balances precision and recall to maximize structural alignment. Table~\ref{tab:addlabel} (c) corroborates this finding where the w/ acc + Jaccard E+R configuration attains the highest overall correctness of 0.487 for reasoning semantic fidelity. Finally, Figure~\ref{fig:reward_traj} illustrates the training trajectories. Unlike the volatile Answer reward, the Entity-Relation Jaccard scores show a consistent upward trend, confirming the stability of our process supervision. Notably, while Precision improves rapidly before plateauing, Recall proves more challenging, stagnating or even slightly declining in later stages.

\begin{figure*}[t!]
\centering
\begin{minipage}[b]{0.49\textwidth}
    \centering
    \includegraphics[width=\textwidth]{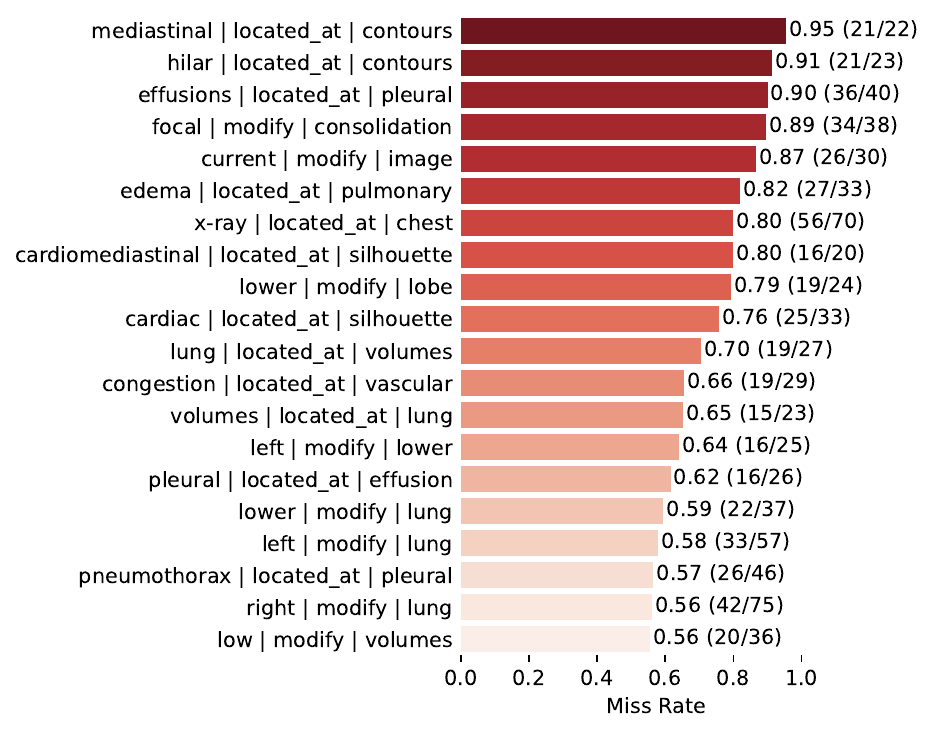}
    \centerline{\small \textbf{(a)} }
\end{minipage}
\hfill
\begin{minipage}[b]{0.49\textwidth}
    \centering
    \includegraphics[width=\textwidth]{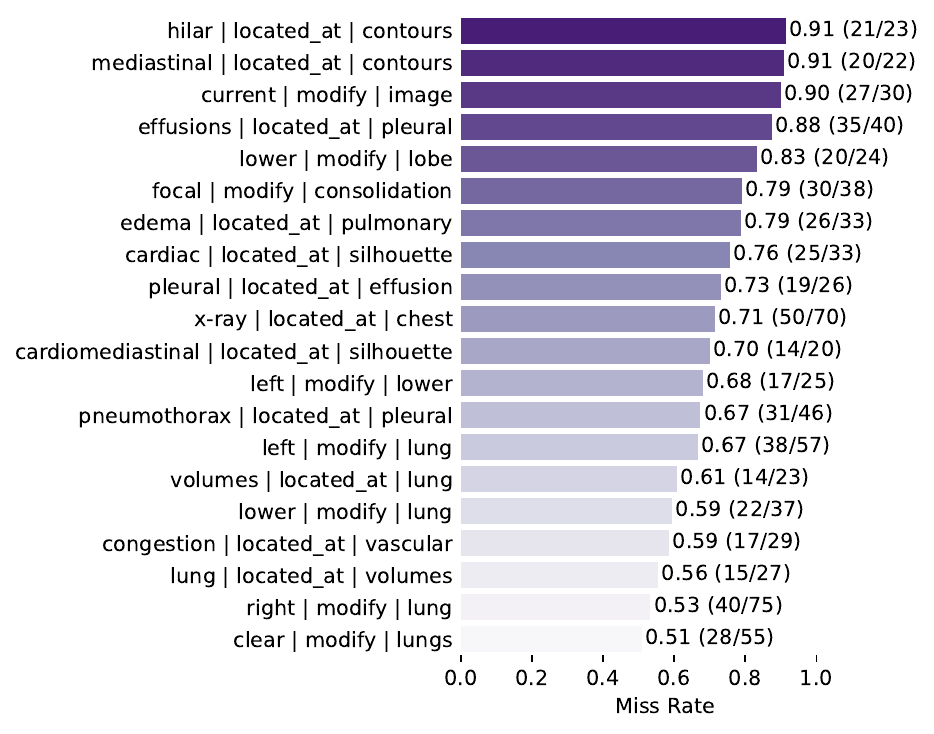}
    \centerline{\small \textbf{(b)} }
\end{minipage}
\caption{\textbf{Miss rates of relations with frequency $>20$, sorted from high to low.} We compare the performance of (a) the baseline SFT model and (b) the proposed CheXPO-v2 model.}
\label{fig:kg_missrate}
\end{figure*}

\begin{figure*}[h!]
    \centering
    \includegraphics[width=1\linewidth]{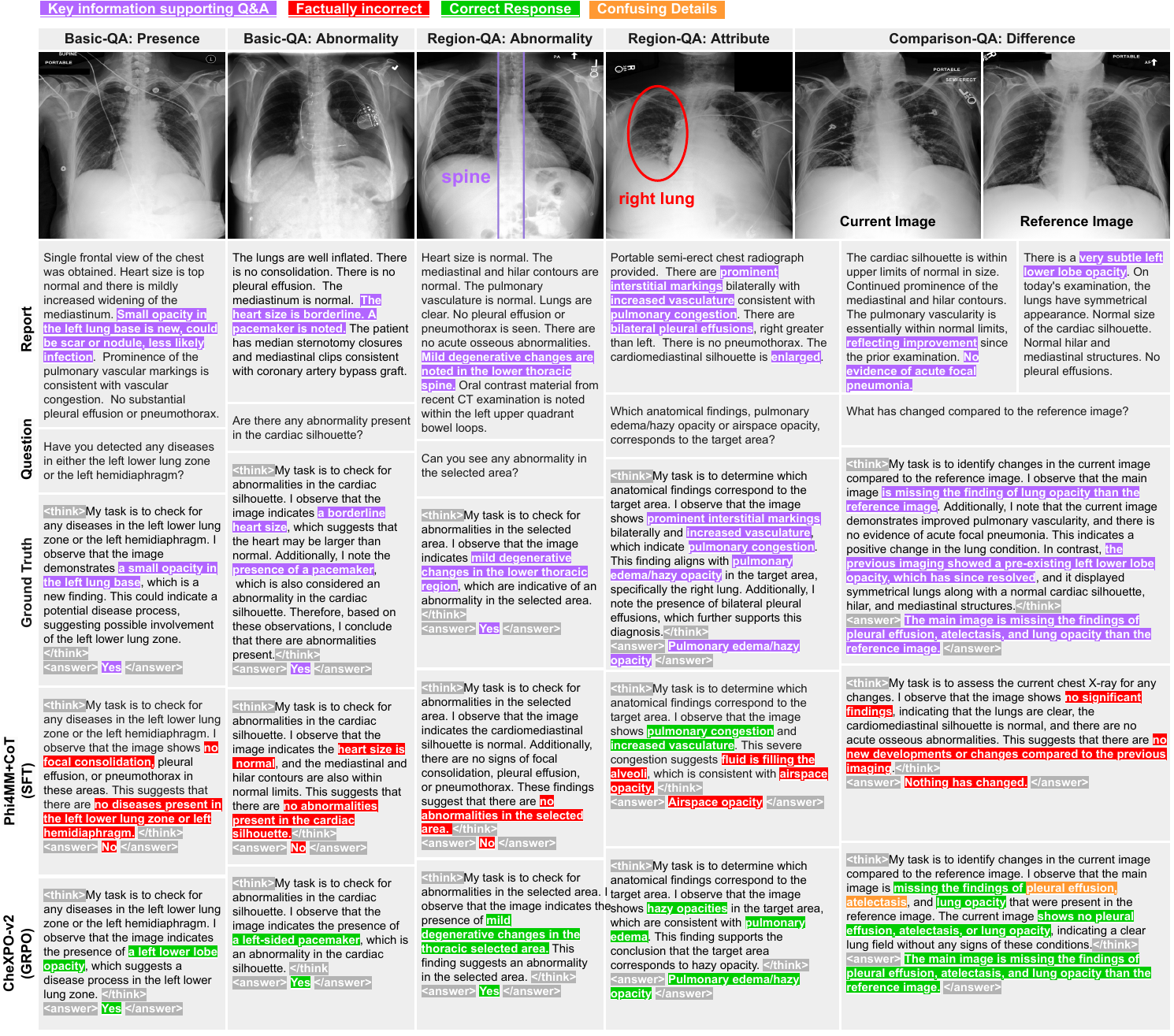}
    \vspace{-0.1cm}
    \caption{Representative chest X-ray cases across three major tasks (Basic-QA, Region-QA, Comparison-QA), comparing outputs from our SFT, and CheXPO-v2 models. }
    \label{fig:caser}
\end{figure*}

\subsection{Qualitative Analysis}
\subsubsection{Error Distribution Analysis}
\label{sec:error_analysis}
Figure \ref{fig:kg_missrate} visualizes the miss rates for high-frequency relation triplets (frequency $>20$) to diagnose specific weaknesses in medical fact recall. In the \textbf{SFT baseline} (a), we observe consistently high miss rates across complex anatomical and pathological descriptions, particularly for abstract structural relations such as ``\textit{mediastinal $|$ located\_at $|$ contours}'' (Miss Rate: 0.95) and ``\textit{effusions $|$ located\_at $|$ pleural}'' (0.90). This suggests that the SFT model often omits necessary relational context. In contrast, our \textbf{CheXPO-v2} model (b) demonstrates a general reduction in miss rates across most categories. For example, the miss rate for ``\textit{mediastinal $|$ located\_at $|$ contours}'' decreases to 0.91, and ``\textit{effusions $|$ located\_at $|$ pleural}'' drops to 0.88. Further improvements are observed in fundamental grounding relations; for instance, the miss rate for ``\textit{x-ray $|$ located\_at $|$ chest}'' decreases from 0.80 to 0.71, and ``\textit{congestion $|$ located\_at $|$ vascular}'' improves from 0.66 to 0.59. This quantitative shift confirms that our knowledge graph consistency reward effectively penalizes the omission of critical medical facts, forcing the model to generate more complete and structurally rigorous reasoning chains.

\subsubsection{Visual Case Visualization}
\label{sec:visual_cases}

Figure \ref{fig:caser} presents representative qualitative comparisons across three major chest X-ray VQA tasks: Basic-QA, Region-QA, and Comparison-QA. We compare the baseline \textbf{SFT} model (Phi4MM+CoT) against our proposed \textbf{CheXPO-v2} (GRPO) model.
The visual analysis highlights a critical limitation in the SFT baseline: it frequently suffers from hallucinations and a ``\textit{normality bias},'' often generating plausible-sounding but factually incorrect statements (marked in \textbf{\textcolor{red}{red}}). For instance, in the \textit{Basic-QA} examples, the SFT model fails to detect obvious abnormalities, incorrectly claiming ``\textit{no focal consolidation}'' or ``\textit{no abnormalities}'' despite the presence of a \textbf{left lower lobe opacity} and a \textbf{pacemaker}. Similarly, in the \textit{Comparison-QA} task, the SFT model erroneously concludes that ``\textit{nothing has changed},'' missing significant disease progression.
In contrast, \textbf{CheXPO-v2} demonstrates superior reasoning fidelity (marked in \textbf{\textcolor{green}{green}}). By leveraging the knowledge graph consistency reward, it accurately identifies fine-grained anatomical and pathological details, such as the ``\textit{left-sided pacemaker}'' and ``\textit{mild degenerative changes}'' in the thoracic spine. Crucially, in the comparison task, CheXPO-v2 correctly reasons that the current image is ``\textit{missing the findings of pleural effusion... and lung opacity},'' effectively capturing the improvement in the patient's condition. This confirms that our method not only improves answer accuracy but also fosters clinically verifiable reasoning processes.

\section{Conclusion}
In this paper, we introduced \textbf{CheXPO-v2}, a verifiable reinforcement learning framework designed to mitigate hallucinations in medical VLMs. By shifting from sparse outcome rewards to fine-grained process supervision via Entity-Relation Matching, our approach effectively penalizes ``overthinking'' and enforces logically sound reasoning chains. Extensive experiments demonstrate that CheXPO-v2 achieves state-of-the-art diagnostic accuracy and reasoning fidelity with exceptional data efficiency, requiring only 5k samples. Future work will focus on enhancing visual interpretability, specifically by integrating region-aware mechanisms to tightly couple textual reasoning with fine-grained visual evidence.



\bibliographystyle{IEEEtran}
\bibliography{reference}

\end{document}